\documentclass[conference,10pt,phv]{IEEEtran}
\usepackage{cite}
\usepackage{amsmath,amssymb,amsfonts}
\usepackage{algorithmic}
\usepackage{graphicx}
\usepackage{siunitx}
\graphicspath{{artwork/}} 
\usepackage{textcomp}
\usepackage{xcolor}
\usepackage{fancyhdr}

\def\BibTeX{{\rm B\kern-.05em{\sc i\kern-.025em b}\kern-.08em
    T\kern-.1667em\lower.7ex\hbox{E}\kern-.125emX}}

\usepackage{booktabs}
\usepackage[
pdfauthor={Tim Prangemeier},hidelinks, bookmarks={false}]{hyperref}

\usepackage{soul}
\usepackage[outline]{contour}
\usepackage{mathtools}
\usepackage{tikz}
\usetikzlibrary{shapes.arrows}
\usetikzlibrary{arrows.meta}
\usepackage{pgfplots}
\usetikzlibrary{patterns}
\usetikzlibrary{shapes.arrows}
\usepackage{tabularx}
\usepackage{dsfont}
\usepackage{threeparttable}
\definecolor{tab:TUred9a}{RGB}{233, 80, 62}
\definecolor{tab:TUgreen3a}{RGB}{80, 182, 149}
\definecolor{tab:TUblue1a}{RGB}{93, 133, 195}
\definecolor{tab:tud0c}{cmyk/RGB/HTML}{0,0,0,.6/137,137,137/898989}

\newcommand{\fakesubsubsection}[1]{%
	\par\refstepcounter{subsubsection}
	\subsectionmark{#1}
	\addcontentsline{toc}{subsubsection}{\protect\numberline{\thesubsection}#1}
}

\definecolor{tab:trapgr1}{RGB}{76, 76, 76}
\definecolor{tab:trapgr2}{RGB}{127, 127, 127}

\definecolor{tab:darkviolet}{RGB}{148, 0, 211}
\definecolor{tab:bggrey}{RGB}{163, 163, 163}
\definecolor{tab:trapblack}{RGB}{0, 0, 0}
\definecolor{tab:hotpink}{RGB}{255, 105, 180}
\definecolor{tab:plum}{RGB}{221, 160, 221}
\definecolor{tab:violet}{RGB}{238, 130, 238}
\definecolor{tab:detr_violet}{RGB}{255,127,231}
\definecolor{tab:trap_grey}{RGB}{86,86,86}
\newcommand \colorindicator[2]{%
	\begingroup%
	\setul{0.25ex}{0.4ex}%
	\contourlength{0.2ex}%
	\setulcolor{#1}%
	{{\phantom{#2}}}\llap{\contour{white}{#2}} \textcolor{#1}{\tiny{$\blacksquare \hspace{-.5mm} \blacksquare$}}%
	\endgroup%
}

\usepackage[colorinlistoftodos]{todonotes}

\newcommand{\tds}[3][]{
	\todo[color=gray!40,inline,#1,caption={#2},disable] {\underline{#2:} \\#3} }

\newcommand{\tdsCamera}[3][]{
	\todo[color=gray!40,inline,#1,caption={Camera Ready: #2},nolist,disable] {\underline{Camera Ready To Do:}\\ #2: \\#3} }

\definecolor{tud0d}{cmyk/RGB/HTML}{0,0,0,.8/83,83,83/535353}
\definecolor{tud0c}{cmyk/RGB/HTML}{0,0,0,.6/137,137,137/898989}
\definecolor{tud0b}{cmyk/RGB/HTML}{0,0,0,.4/181,181,181/B5B5B5}
\definecolor{tud0a}{cmyk/RGB/HTML}{0,0,0,.2/220,220,220/DCDCDC}
\definecolor{tud1a}{cmyk/RGB/HTML}{.7,.4,0,0/93,133,195/5D85C3}
\definecolor{tud2a}{cmyk/RGB/HTML}{0.8,.2,0,0/0,156,218/009CDA}
\definecolor{tud3a}{cmyk/RGB/HTML}{0.7,0,.5,0/80,182,149/50B695}
\definecolor{tud4a}{cmyk/RGB/HTML}{.4,0,.8,0/175,204,80/AFCC50}
\definecolor{tud5a}{cmyk/RGB/HTML}{.2,0,.8,0/221,223,72/DDDF48}
\definecolor{tud6a}{cmyk/RGB/HTML}{0,.1,.7,0/255,224,92/FFE05C}
\definecolor{tud7a}{cmyk/RGB/HTML}{0,.3,.8,0/248,186,60/F8BA3C}
\definecolor{tud8a}{cmyk/RGB/HTML}{0,.6,.8,0 /238,122,52/EE7A34}
\definecolor{tud9a}{cmyk/RGB/HTML}{0,.8,.7,0/233,80,62/E9503E}
\definecolor{tud10a}{cmyk/RGB/HTML}{.2,.9,0,0/201,48,142/C9308E}
\definecolor{tud11a}{cmyk/RGB/HTML}{.6,.8,0,0/128,69,151/804597}
\definecolor{tud1b}{cmyk/RGB/HTML}{1,.6,0,0/0,90,169/005AA9}
\definecolor{tud2b}{cmyk/RGB/HTML}{1,.3,0,0/0,131,204/0083CC}
\definecolor{tud3b}{cmyk/RGB/HTML}{1,0,.6,0/0,157,129/009D81}
\definecolor{tud4b}{cmyk/RGB/HTML}{.5,0,1,0/153,192,0/99C000}
\definecolor{tud5b}{cmyk/RGB/HTML}{.3,0,1,0/201,212,0/C9D400}
\definecolor{tud6b}{cmyk/RGB/HTML}{0,.2,1,0/253,202,0/FDCA00}
\definecolor{tud7b}{cmyk/RGB/HTML}{0,.4,1,0/245,163,0/F5A300}
\definecolor{tud8b}{cmyk/RGB/HTML}{0,.7,1,0/236,101,0/EC6500}
\definecolor{tud9b}{cmyk/RGB/HTML}{0,1,.9,0/230,0,26/E6001A}
\definecolor{tud10b}{cmyk/RGB/HTML}{.4,1,0,0/166,0,132/A60084}
\definecolor{tud11b}{cmyk/RGB/HTML}{.7,1,0,0/114,16,133/721085}
\definecolor{tud1c}{cmyk/RGB/HTML}{1,.7,.2,0/0,78,138/004E8A}
\definecolor{tud2c}{cmyk/RGB/HTML}{1,.5,.2,0/0,104,157/00689D}
\definecolor{tud3c}{cmyk/RGB/HTML}{1,.2,.6,0/0,136,119/008877}
\definecolor{tud4c}{cmyk/RGB/HTML}{.6,.1,1,0/127,171,22/7FAB16}
\definecolor{tud5c}{cmyk/RGB/HTML}{.4,.1,1,0/177,189,0/B1BD00}
\definecolor{tud6c}{cmyk/RGB/HTML}{.2,.3,1,0/215,172,0/D7AC00}
\definecolor{tud7c}{cmyk/RGB/HTML}{.2,.5,1,0/210,135,0/D28700}
\definecolor{tud8c}{cmyk/RGB/HTML}{.2,.8,1,0/204,76,3/CC4C03}
\definecolor{tud9c}{cmyk/RGB/HTML}{.3,1,.9,0/185,15,34/B90F22}
\definecolor{tud10c}{cmyk/RGB/HTML}{.5,1,.3,0/149,17,105/951169}
\definecolor{tud11c}{cmyk/RGB/HTML}{.8,1,.2,0/97,28,115/611C73}
\definecolor{tud1d}{cmyk/RGB/HTML}{1,.9,.3,0/36,53,114/243572}
\definecolor{tud2d}{cmyk/RGB/HTML}{1,.7,.4,0/0,78,115/004E73}
\definecolor{tud3d}{cmyk/RGB/HTML}{1,.4,.7,0/0,113,94/00715E}
\definecolor{tud4d}{cmyk/RGB/HTML}{.7,.3,1,0/106,139,55/6A8B22}
\definecolor{tud5d}{cmyk/RGB/HTML}{.5,.2,1,0/153,166,4/99A604}
\definecolor{tud6d}{cmyk/RGB/HTML}{.4,.4,1,0/174,142,0/AE8E00}
\definecolor{tud7d}{cmyk/RGB/HTML}{.3,.6,1,0/190,111,0/BE6F00}
\definecolor{tud8d}{cmyk/RGB/HTML}{.4,.8,1,0/169,73,19/A94913}
\definecolor{tud9d}{cmyk/RGB/HTML}{.5,1,.9,0/156,28,38/961C26}
\definecolor{tud10d}{cmyk/RGB/HTML}{.7,1,.5,0/115,32,84/732054}
\definecolor{tud11d}{cmyk/RGB/HTML}{.9,1,.3,0/76,34,106/4C226A}



\makeatletter
\newcommand*\titleheader[1]{\gdef\@titleheader{#1}}
\AtBeginDocument{%
	\let\st@red@title\@title%
	\def\@title{%
		\bgroup\normalfont\normalsize\centering\@titleheader\par\egroup
		\vskip0.5em\st@red@title}
} 
\makeatother

\title{Attention-Based Transformers for Instance Segmentation of Cells in Microstructures}
\titleheader{\vspace{-11mm} 2020 IEEE International Conference on Bioinformatics and Biomedicine (BIBM)}

\begin{document}
\pagenumbering{roman}


\author{ 
\IEEEauthorblockN{
	Tim Prangemeier, 
	Christoph Reich, 
	Heinz Koeppl\IEEEauthorrefmark{3}}
\vspace{0.05in}\IEEEauthorblockA{Centre for Synthetic Biology,\\Department of Electrical Engineering and Information Technology, Department of Biology,\\
	Technische Universit\"at Darmstadt\\
	\emph{\IEEEauthorrefmark{3}heinz.koeppl@bcs.tu-darmstadt.de}}
}

%
%

\maketitle
\thispagestyle{plain}
\fancypagestyle{plain}{
	\fancyhf{} 
	\fancyfoot[L]{978-1-7281-6215-7/20/\$31.00~\copyright2020~IEEE} 
	\renewcommand{\headrulewidth}{0pt}
	\renewcommand{\footrulewidth}{0pt}
}

\begin{abstract}
Detecting and segmenting object instances is a common task in biomedical applications. Examples range from detecting lesions on functional magnetic resonance images, to the detection of tumours in histopathological images and extracting quantitative single-cell information from microscopy imagery, where cell segmentation is a major bottleneck. Attention-based transformers are state-of-the-art in a range of deep learning fields. They have recently been proposed for segmentation tasks where they are beginning to outperform other  methods. We present a novel attention-based \textit{cell detection transformer} (Cell-DETR) for direct end-to-end instance segmentation. While the segmentation performance is on par with a state-of-the-art instance segmentation method, Cell-DETR is simpler and faster. We showcase the method's contribution in a the typical use case of segmenting yeast in microstructured environments, commonly employed in systems or synthetic biology. For the specific use case, the proposed method surpasses the state-of-the-art tools for semantic segmentation and additionally predicts the individual object instances. The fast and accurate instance segmentation performance increases the experimental information yield for \emph{a posteriori} data processing and  makes online monitoring of experiments and closed-loop optimal experimental design feasible.\\
Code and data samples are available at  \url{https://git.rwth-aachen.de/bcs/projects/cell-detr.git}.
\tds{novel? omit novel}{}

\end{abstract}



\section{Introduction}
\label{sec:introduction}
\begin{figure}[htbp]
	\centerline
	{\input{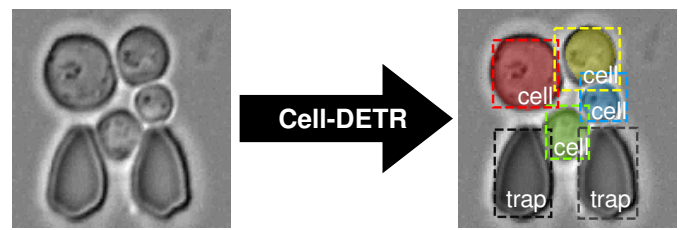}}
	\vspace{-2mm}
	\caption[$\blacksquare$ Fig. \ref{fig:intro_schem}: simple intro schematic  DETR (\checkmark)]{Schematic of Cell-DETR direct instance segmentation discerning individual cell (colour) and trap microstructure (grey) object instances.}
	\label{fig:intro_schem}
\end{figure}

\addcontentsline{tdo}{todo}{\textbf{Introduction}}
\setcounter{page}{1}
\pagenumbering{arabic}
\fakesubsubsection{\P \  instance seg to biomed application (\checkmark)}
\label{sssec:instseg}
\addcontentsline{toc}{subsubsection}{\nameref{sssec:instseg}}
Instance segmentation is a common task in biomedical applications. It is comprised of both detecting individual object instances and segmenting them \cite{He2017,Cordts2016}. Prevalent examples in healthcare and life sciences include the detection of individual tumour or cell entities and the segmentation of their shape. Recent advances in automated single-cell image processing, such as instance segmentation, have contributed to early tumour detection, personalised medicine, biological signal transduction and insight into the mechanisms behind cell population heterogeneity, amongst others \cite{Sun2020,Leygeber2019,Hofmann2019,Prangemeier2020b}. An example of  instance segmentation is shown in Fig. \ref{fig:intro_schem}, with four separate cell and two trap microstructures detected and segmented individually. 
\fakesubsubsection{\P \ attention-based methods DETR (\checkmark)}
\label{sssec:attentro}
\addcontentsline{toc}{subsubsection}{\nameref{sssec:attentro}}
Object detection and panoptic segmentation are closely related to instance segmentation \cite{Kirillov2019,Cordts2016}. Carion \textit {et al.} recently proposed a novel attention-based \textit{detection transformer} DETR  for panoptic segmentation \cite{Carion2020}. DETR achieves state-of-the-art panoptic segmentation performance, while exhibiting a comparatively simple architecture that is easier to implement and is computationally more efficient than previous approaches \cite{Carion2020}. Its simplicity  promises to be beneficial for its adoption in real-world applications. 
\fakesubsubsection{\P \ TLFM synbio (\checkmark)}
\label{sssec:tlfm-synbio}
\addcontentsline{toc}{subsubsection}{\nameref{sssec:tlfm-synbio}}
Time-lapse fluorescence microscopy (TLFM) is a powerful technique for studying cellular processes in living cells \cite{Leygeber2019,Lugagne2020,Pepperkok2006,Prangemeier2020}. The vast amount of quantitative data TLFM yields, promises to constitute the backbone of the rational design of \emph{de novo} biomolecular functionality \cite{Pepperkok2006,Prangemeier2020}. Ideally in synthetic biology, well characterised parts are combined \emph{in silico} in a quantitatively predictive bottom up approach \cite{Gomez2019,Lehr2019,Prangemeier2020}, for example, to detect and kill cancer cells \cite{Xie2011,Si2018}. 
\fakesubsubsection{\P \ TLFM image cyto traps $\circlearrowleft$}
\label{sssec:tlfm-exp}
\addcontentsline{toc}{subsubsection}{\nameref{sssec:tlfm-exp}}
Quantitative TLFM with high-throughput microfluidics is an essential technique for concurrently studying the heterogeneity and dynamics of synthetic circuitry on the single cell level \cite{Leygeber2019,Lugagne2020,Prangemeier2020}. A typical TLFM experiment yields thousands of specimen images (Fig. \ref{fig:intro_schem}) requiring automated segmentation, examples include \cite{Hofmann2019,Bakker2017,Crane2014}.  Segmenting each individual cell enables its pertinent information to be extracted quantitatively. For example, the abundance of a fluorescent reporter can be measured, giving insight into the cell's inner workings. 
\fakesubsubsection{\P \ Bottleneck / Problem formulation $\circlearrowleft$}
\label{sssec:bottleneck}
\addcontentsline{toc}{subsubsection}{\nameref{sssec:bottleneck}}
Instance segmentation is a major bottleneck in quantifying single-cell microscopy data and manual analysis is prohibitively labour intensive \cite{Bakker2017,VanValen2016,Sauls2019,Lugagne2020,Prangemeier2020}. The vast majority of single-cell segmentation methods are designed for \textit{a posteriori} data processing and often require post-processing for instance detection or manual input \cite{Lugagne2020}. This is not only a drawback on the amount of experiments that can be performed, but also limits the type of experiments \cite{VanValen2016,Moen2019}. For example, harnessing the potential of advanced closed-loop optimal experimental design techniques \cite{Gomez2019,Prangemeier2018,Bandiera2020} requires online monitoring with fast instance segmentation capabilities. Attention-based methods, such as the recently proposed detection transformer DETR \cite{Carion2020}, are increasingly outperforming other methods \cite{Vaswani2017,Carion2020}. For the yeast-trap configuration  (Fig. \ref{fig:intro_schem}) direct instance segmentation has yet to be employed and attention-based transformers have yet to be applied for segmentation in the biomedical fields in general.
\fakesubsubsection{\P \ This study (\checkmark)}
\label{sssec:here}
\addcontentsline{toc}{subsubsection}{\nameref{sssec:here}}
In this study, we present Cell-DETR, a novel attention-based detection transformer for instance segmentation of biomedical samples based on DETR \cite{Carion2020}. We address the  automated cell instance segmentation bottleneck for yeast cells in microstructured environments (Fig. \ref{fig:intro_schem}) and showcase Cell-DETR on this application. 
Section \ref{sec:background} introduces the  previous segmentation approaches and the microstructured environment. Our experimental setup for fluorescence microscopy, the tested architectures and  our approach to training and evaluation are presented in Section \ref{sec:method}. We analyse the proposed method's performance in Section \ref{sec:results} and compare it to the application specific state-of-the-art, as well as to a general instance segmentation baseline. After interpreting the results and highlighting the method's future potential in Section \ref{sec:discussion}, we summarise and conclude the study in Section \ref{sec:conclusion}. Our model surpasses the previous application baseline and is on par with a general state-of-the-art instance segmentation method. The relatively short inference runtimes enable higher throughput \textit{a posteriori} data processing and make online monitoring of experiments with approximately $1000$ cell traps feasible.

\section{Background }
\label{sec:background}
\addcontentsline{tdo}{todo}{\textbf{Background}}
\fakesubsubsection{\P \ Trapped cell segmentation tools (\checkmark)}
\label{sssec:single_seg}
\addcontentsline{toc}{subsubsection}{\nameref{sssec:single_seg}}
An extensive body of research into the automated processing of microscopy imagery dates back to the middle of the 20-th century. Recent studies demonstrate the utility of deep learning segmentation approaches, for example \cite{Ronneberger2015,Lugagne2020,Sauls2019,Prangemeier2020b,Dietler2020}. Comprehensive reviews of the many methods to segment yeast on microscopy imagery are available elsewhere \cite{Moen2019,Sun2020}. Here we focus on dedicated tools for segmenting cells in trapped microstructures. U-Net convolutional neural networks (CNNs) with an encoder-decoder architecture bridged by skip connections have been shown to perform semantic segmentation well for \textit{E. coli} mother machines \cite{Lugagne2020,Sauls2019} and yeast in microstructured environments \cite{Prangemeier2020b}. In the case of trapped yeast, the previous state-of-the-art tool DISCO \cite{Bakker2017} was based on conventional methods (template matching, support vector machine, active contours), until recently being superseded by U-Nets \cite{Prangemeier2020b}. The current baseline for semantic segmentation of yeast in microstructured environments, as measured by the cell class intersection-over-union, is $0.82$\cite{Prangemeier2020b}. Additional post-processing, of the segmentation maps is required to attain each individual cell instance \cite{Prangemeier2020b}. 
\begin{figure}[htbp]
	\centerline
	{\includegraphics[width=1\columnwidth]{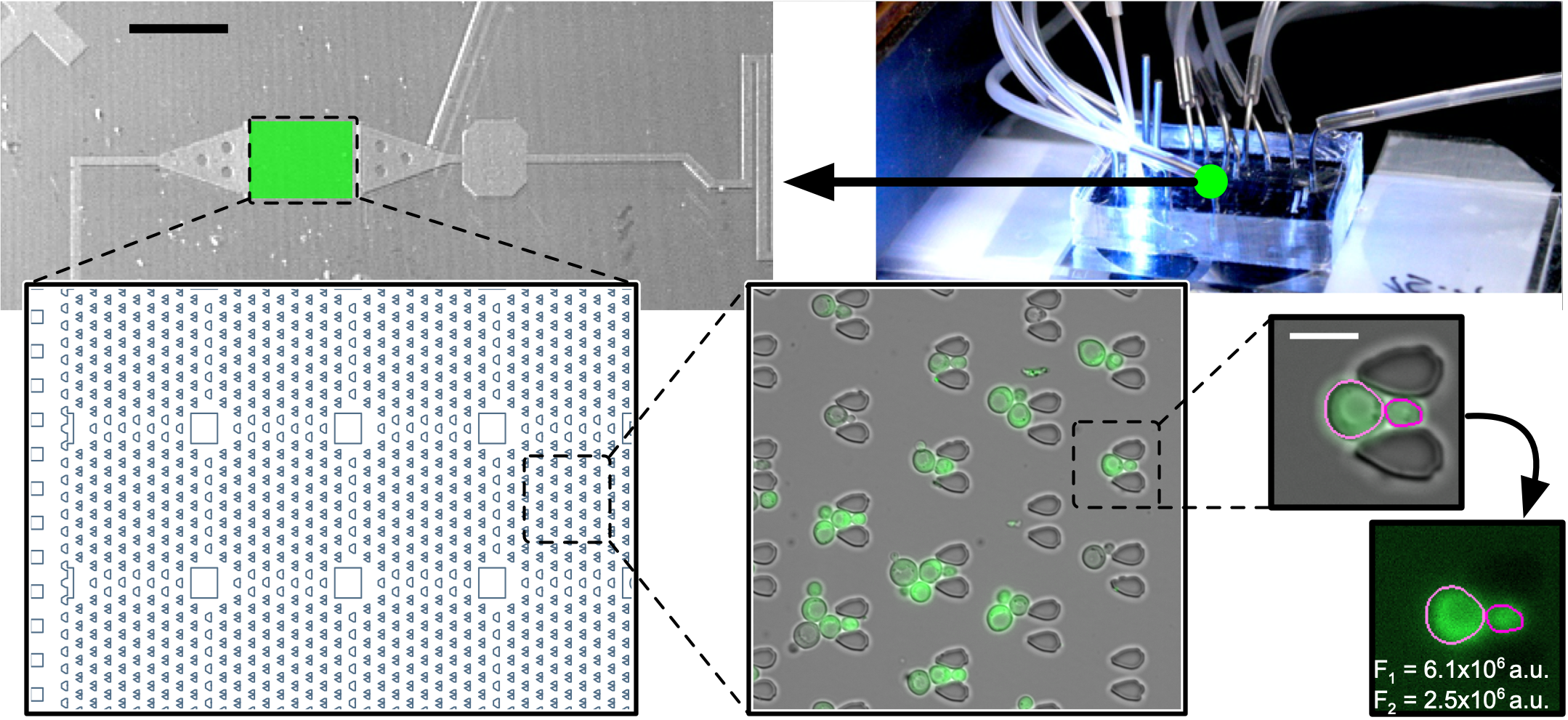}}
	\vspace{-2mm}
	\caption[$\square$ Fig. \ref{fig:complex}: complex whole chip $\circlearrowleft$]{Single-cell fluorescence measurement setup. Microfluidic chip on the microscope table (top right), microscope imagery and design of the yeast trap microstructures. The trap chamber (green rectangle) contains an array of approximately \SI{1000}{} traps. Single specimen images show a pair of microstructures and  fluorescent cells, violet contours indicate segmentation of two separate cell instances with corresponding fluorescence measurement $F_1$ and $F_2$; black scale bar \SI{1}{\milli\meter}, white scale bar \SI{10}{\micro\meter}.}
	\label{fig:complex}
\end{figure}
\fakesubsubsection{\P \  recent, more advanced methods to DETR}
\label{sssec:advanced}
\addcontentsline{toc}{subsubsection}{\nameref{sssec:advanced}}
For instance segmentation in general, recent state-of-the-art methods are available, for example Mask R-CNN \cite{He2017}. It is a proposal-based instance segmentation model, which combines a CNN backbone, region proposals with non-maximum-suppression, region-of-interest (ROI) pooling, and multiple prediction heads \cite{He2017}. Attention-based methods are increasingly outperforming convolutional methods and are currently state-of-the-art in natural language processing \cite{Vaswani2017}. Beyond natural language processing, attention-based approaches, such as axial-attention modules \cite{Wang2020}, have demonstrated promising results in computer vision applications \cite{Carion2020}. Recently, the first transformer-based method (DETR \cite{Carion2020}) for object detection and panoptic segmentation was reported. DETR achieves state-of-the-art results on par with Faster R-CNN and constitutes a promising approach for further improvements in automated object detection and segmentation performance.
\fakesubsubsection{\P \ Microfluidically trapped yeast  culture $\circlearrowleft$}
\label{sssec:image_cytometry}
\addcontentsline{toc}{subsubsection}{\nameref{sssec:image_cytometry}}
The microfluidic trap microstructures we consider here are designed for long-term culture of yeast cells (\emph{Saccharomyces cerevisiae}) within the focal plane of a microscope \cite{Crane2014}. The on-chip environment is tightly controlled and conducive to yeast growth. Examples of its routine employ include Fig. \ref{fig:complex} and \cite{Prangemeier2020b,Prangemeier2020,Hofmann2019,Bakker2017,Leygeber2019}. A constant flow of yeast growth media hydrodynamically traps the cells in the microstructures and allows the introduction of chemical perturbations. An automated microscope records an entire trap chamber of up to $1000$ traps by imaging both the brightfield and fluorescent channels at approximately \SI{20}{} neighbouring positions. Typical experiments each produce hundreds of GB of image data. Time-lapse recordings allow individual cells to be tracked through time. Robust instance segmentation facilitates tracking \cite{Moen2019,Lugagne2020}, which itself can be a limiting factor with regard to the data yield of an experiment \cite{Moen2019,Leygeber2019,Sauls2019,Lugagne2020}.


\section{Methodology}
\label{sec:method}
\addcontentsline{tdo}{todo}{\textbf{Methodology}}

\subsection{Live-cell microscopy dataset and annotations}
\label{ssec:data}
\addcontentsline{tdo}{subsection}{\nameref{ssec:data}}

\fakesubsubsection{\P \ Specimen images \checkmark}
\label{sssec:images}
\addcontentsline{toc}{subsubsection}{\nameref{sssec:images}}

\begin{figure}[htbp]
	\centerline
	{\includegraphics{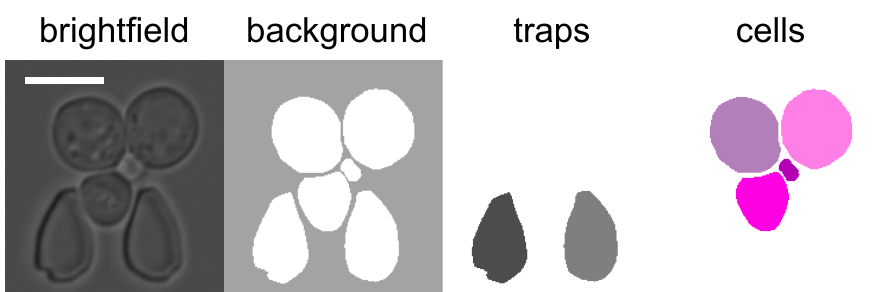}}
		\vspace{-2mm}
	\caption[$\blacksquare$ Fig. \ref{fig:classes_def}: classes (\checkmark)]{Example of class and instance annotations for a specimen image; brightfield image (left), background label in \colorindicator{tab:bggrey}{light grey}, instances of the trap class in shades of  \colorindicator{tab:trapgr1}{dark grey} and instances of the cell class in shades of  \colorindicator{tab:detr_violet}{violet} (left to right respectively); scale bar \SI{10}{\mu \meter}.}
	\label{fig:classes_def}
\end{figure}

\begin{figure}[htbp]
	\centerline
	{\includegraphics{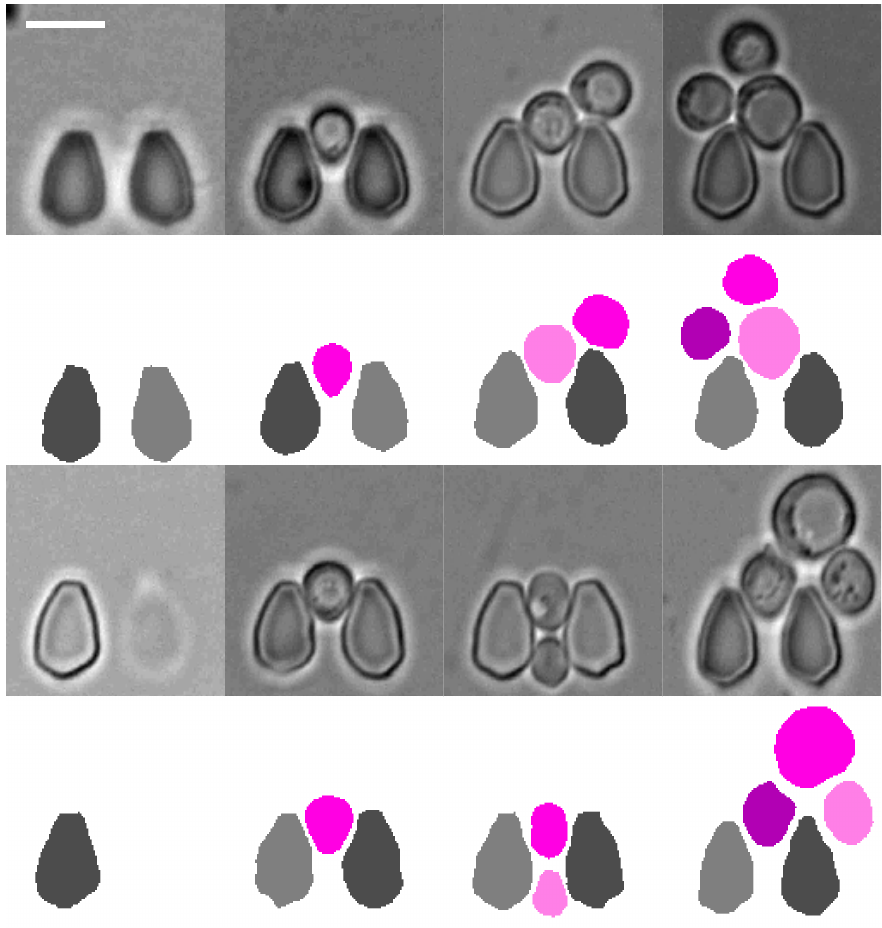}}
	\vspace{-2mm}
	\caption[$\blacksquare$ Fig. \ref{fig:tradata}: training data (\checkmark)]{Characteristic selection of specimen images and corresponding annotations, including empty or single trap structures, trapped single cells (with single daughter adjacent) and multiple trapped cells; trap instances in shades of \colorindicator{tab:trap_grey}{dark grey}, cell instances in shades of \colorindicator{tab:detr_violet}{violet} and transparent background; scale bar \SI{10}{\mu \meter}.}
	\label{fig:tradata}
\end{figure}

The individual specimen images each contain a single microfluidic trap and some yeast cells, as depicted in Fig. \ref{fig:intro_schem}. These are extracted from larger microscope recordings, whereby each exposure contains up to \si{50} traps (Fig. \ref{fig:complex} middle). Ideally, a single mother cell persists in each trap, with subsequent daughter cells being removed by the constant flow of yeast growth media. In practice, multiple cells accumulate around some traps, while other traps remain empty (Fig. \ref{fig:classes_def}).
\tds{add sentence on omitting the bounding boxes on imagery for better view of the contours, indicate instances by shading}{}
\fakesubsubsection{\P \ Classes annotations \checkmark}
\label{sssec:classes}
\addcontentsline{toc}{subsubsection}{\nameref{sssec:classes}}
We distinguish between three classes on the specimen image annotations, as depicted in Fig. \ref{fig:classes_def}. The yeast cells in violet are the most important class for biological applications. To counteract traps being segmented as cells, we employ a distinct class for them (dark grey). The background (light grey) is annotated for semantic segmentation training, for example of U-Nets.  For instance segmentation training we introduce a \textit{no-object} class $\varnothing$ in place of the background class. 
\fakesubsubsection{\P \ Object instance annotations \checkmark}
\label{sssec:object_instances}
\addcontentsline{toc}{subsubsection}{\nameref{sssec:object_instances}}
Each instance of cells or trap structures are annotated individually with a bounding box, class specification and separate pixel-wise segmentation map. Here we omit the bounding boxes to enable an unobscured view of the contours. Instead, the distinct cell instances and their individual segmentation maps are indicated by different shades of violet in Fig. \ref{fig:classes_def}. 
\fakesubsubsection{\P \ Training, validation and test data \checkmark}
\label{sssec:tradata}
\addcontentsline{toc}{subsubsection}{\nameref{sssec:tradata}}
The annotated set of $419$ specimen images from various experiments was randomly assigned for network training, validation and testing (\SI{76}{\percent}, \SI{12}{\percent} and \SI{12}{\percent} respectively). Examples are shown in Fig. \ref{fig:tradata}. Images include a balance of the common yeast-trap configurations: 1) empty traps, 2) single cells (with daughter) and 3) multiple cells. Slight variations in trap fabrication, debris, contamination, focal shift, illumination levels and yeast morphology were included. Further scenarios or strong variations, such as trap design geometries, model organisms and significant focal shift, were omitted.
\subsection{The Cell-DETR instance segmentation architecture}
\label{ssec:DETR_architecture}

\begin{figure*}[htbp]
	\centerline
	{\input{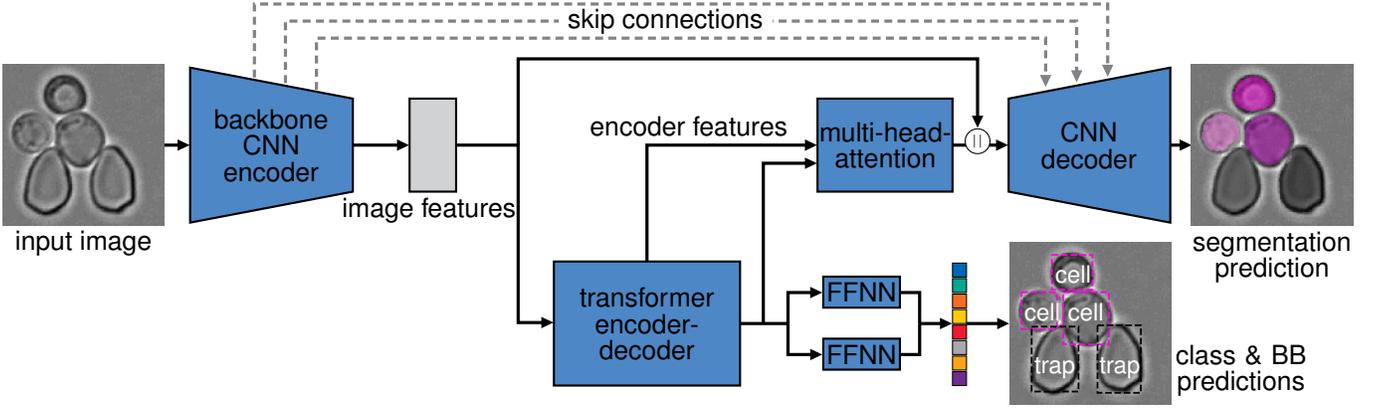}}
	\vspace{-2mm}
	\caption[$\blacksquare$ Fig. \ref{fig:architecture} DETR architecture (\checkmark)]{Architecture of the end-to-end instance segmentation network, with brightfield specimen image input and an instance segmentation prediction as output. The backbone CNN encoder extracts image features that then feed into both the transformer encoder-decoder for class and bounding box prediction, as well as to the CNN decoder for segmentation. The transformer encoded features, as well as the transformer decoded features, are fed into a multi-head-attention module and together with the image features from the CNN backbone feed into the CNN decoder for segmentation. Skip connections additionally bridge between the backbone CNN encoder and the CNN decoder. Input and output resolution is 128 × 128 pixels.}
	\label{fig:architecture}
\end{figure*}

\fakesubsubsection{\P Architecture intro (\checkmark)}
\label{sssec:architecture1}
\addcontentsline{toc}{subsubsection}{\nameref{sssec:architecture1}}
The proposed Cell-DETR models A and B are based on the DETR panoptic segmentation architecture \cite{Carion2020}. We adapted the architecture for non-overlapping instance segmentation and reduced it in size for faster inference. The main differences between DETR and our variants Cell-DETR A and B are summarised in Table \ref{tab:architecture}. The Cell-DETR variants have approximately one order of  magnitude less parameters than the original ($\sim40\times 10^6$ reduced to $\sim5\times10^6$ parameters). The main building blocks of the Cell-DETR model are detailed in Fig. \ref{fig:architecture}. They are the backbone CNN encoder, the transformer encoder-decoder, the bounding box and class prediction heads, and the segmentation head. 
\fakesubsubsection{\P Architecture 2: backbone (\checkmark)}
\label{sssec:architecture2}
\addcontentsline{toc}{subsubsection}{\nameref{sssec:architecture2}}
The CNN encoder (left in Fig. \ref{fig:architecture}) extracts image features of the brightfield specimen image input. It is based on four ResNet-like \cite{He2016} blocks with $64$, $128$, $256$ and $256$ convolutional filters. After each block a $2\times 2$ average pooling layer is utilised to downsample the intermediate feature maps. The Cell-DETR variants employ different activations and convolutions, as detailed in Table \ref{tab:architecture}. 
\fakesubsubsection{\P Architecture 3: TR encoder decoder (\checkmark)}
\label{sssec:architecture3}
\addcontentsline{toc}{subsubsection}{\nameref{sssec:architecture3}}
The transformer encoder determines the attention between image features. The transformer decoder predicts the attention regions for each of the $N = 20$ object queries. They are both based on the DETR architecture \cite{Carion2020}. We reduced the number of transformer encoder blocks to three and decoder blocks to two, each with $512$ hidden features in the feed-forward neural-network (FFNN). The $128$ backbone features are flattened before being fed into the transformer. In contrast to the original DETR, we employed learned positional encodings. While Cell-DETR A employs leaky ReLU \cite{Maas2013} activations, Pad\'{e} activation units \cite{Molina2019} are utilised for Cell-DETR B.
\begin{table}[htbp]
	\centering
	\caption[$\boxminus$ Table \ref{tab:architecture}: arch. DETR, C-DETR A and B (\checkmark) ]{Overview of differences between DETR, Cell-DETR A and B.}
	\vspace{-2mm}
	\begin{tabular}{>{\raggedright\arraybackslash}p{1.4cm} >{\centering\arraybackslash}p{1.4cm} >{\centering\arraybackslash}p{1.4cm} >{\centering\arraybackslash}p{1.4cm} >{\raggedleft \arraybackslash}p{0.8cm}} 
		\toprule
		\multicolumn{1}{p{1.4cm}}{\raggedright \textcolor{white}{-------} Model}&%
		\multicolumn{1}{p{1.4cm}}{\centering Activation functions}&%
		\multicolumn{1}{p{1.4cm}}{Convolutions}&%
		\multicolumn{1}{p{1.4cm}}{\centering Feature fusion}&%
		\multicolumn{1}{p{0.8cm}}{\centering Param. $\times 10^{6}$}\\ 
		\midrule 
		DETR \cite{Carion2020} & ReLU & standard spatial 
		& addition 
		& $\gtrapprox 40$ \\
		& & & & \\
		C-DETR A & leaky ReLU \cite{Maas2013} & standard spatial 
		& addition 
		& $\SI{4.3}{}$ \\
		& & & & \\
		C-DETR B & Pad\'{e} \cite{Molina2019} & deformable (v2) \cite{Zhu2019} 
		& pix.-adapt. conv. \cite{Su2019} & $\SI{5.0}{}$ \\
		\bottomrule
	\end{tabular}
	\label{tab:architecture}
\end{table}
\fakesubsubsection{\P Architecture 4: bb and class heads (\checkmark)}
\label{sssec:architecture4}
\addcontentsline{toc}{subsubsection}{\nameref{sssec:architecture4}}
The prediction heads for the bounding box and classification are each a FFNN. They map the transformer encoder-decoder output to the bounding box and classification prediction. These FFNN process each query in parallel and share parameters over all queries. In addition to the cell and trap classes, the classification head can also predict the no-object class $\varnothing$. 
\fakesubsubsection{\P Architecture 5: seg head (\checkmark)}
\label{sssec:architecture5}
\addcontentsline{toc}{subsubsection}{\nameref{sssec:architecture5}}
The segmentation head is composed of a multi-head attention mechanism and a CNN decoder to predict the segmentation maps for each object instance. We employ the original DETR \cite{Carion2020} two-dimensional multi-head attention mechanism between the transformer encoder and decoder features. The resulting attention maps are concatenated channel-wise onto the image features and fed into the CNN decoder. The three ResNet-like decoder blocks decrease the feature channel size while increasing the spatial dimensions. Long skip connections bridge between the CNN encoder and CNN decoder blocks' respective outputs. The features are fused by element-wise addition in Cell-DETR A and by pixel-adaptive convolutions in Cell-DETR B. A fourth convolutional block incorporates the queries in the feature dimension and returns the original input's spatial dimension for each query. Non-overlapping segmentation is ensured by a softmax over all queries.

\subsection{Training Cell-DETR}
\label{ssec:training}
\fakesubsubsection{\P \ Training approach set labels and predictions (\checkmark) }
\label{sssec:lossset}
\addcontentsline{toc}{subsubsection}{\nameref{sssec:lossset}}
We employ a combined loss function and a direct set prediction to train our Cell-DETR networks end-to-end. The set prediction $\hat{y}=\{\hat{y}_{i}=\{\hat{p}_{i},\hat{b}_{i},\hat{s}_{i}\}\}_{1}^{N=20}$ is comprised of the respective predictions for class probability $\hat{p}_{i}\in\mathbb{R}^K$ (here $K=3$ classes, no-object, trap, cell), bounding box $\hat{b}_{i}\in\mathbb{R}^4$ and segmentation ${\hat{s}}_{i}\in\mathbb{R}^{128\times 128}$ for each of the $N$ queries. We assigned each instance set label $y_{\sigma(i)}$ to the corresponding query set prediction $\hat{y_i}$ with the Hungarian algorithm \cite{Carion2020,Kuhn1955}. The indices $\sigma_{i}$ denote the best matching permutation of labels.
\fakesubsubsection{\P \ Training loss function general (\checkmark)}
\label{sssec:lossgen}
\addcontentsline{toc}{subsubsection}{\nameref{sssec:lossgen}}
The combined loss $\mathcal{L}$ is comprised of a classification loss $\mathcal{L}_p$, a bounding box loss $\mathcal{L}_{b}$, and a segmentation loss $\mathcal{L}_{s}$
\begin{equation*}
\mathcal{L}=\sum_{i=1}^{N}\left( \mathcal{L}_{p} + \mathds{1}_{\{p_i \neq \varnothing\}} \mathcal{L}_{b} + \mathds{1}_{\{p_i \neq \varnothing\}} \mathcal{L}_{s}\right),
\end{equation*}
with $N = 20$ object instance queries in this case. We employ class-wise weighted cross entropy for the classification loss
\begin{equation*}
\mathcal{L}_{p}\left(p_{\sigma\left(i\right)}, \hat{p}_{i}\right) = -\sum_{k=1}^{K}\beta_{k}\,p_{\sigma\left(i\right),k}\log(\hat{p}_{i,k}),
\end{equation*}
with weights $\mathbf{\beta}=\left[0.5, 0.5, 1.5\right]$ for the $K = 3$ classes, no-object, trap and cell classes respectively. 
The bounding box loss is itself composed of two weighted loss terms. These are a \textit{generalised intersection-over-union} $\mathcal{L_J}$ \cite{Rezatofighi2019}, and a L1 loss, with respective weights $\lambda_{{\mathcal{J}}}=0.4$ and $\lambda_{{L1}}=0.6$ 
\begin{equation*}
\mathcal{L}_{b}\left(b_{\sigma\left(i\right)}, \hat{b}_{i}\right)=\lambda_{\mathcal{J}}\, \mathcal{L}_\mathcal{J}\left(b_{\sigma\left(i\right)}, \hat{b}_{i}\right)+\lambda_{{L1}}\left|\left| b_{\sigma\left(i\right)}-\hat{b}_{i}\right|\right|_{1}.
\end{equation*}
\fakesubsubsection{\P \ Training loss segmentation (\checkmark)}\label{sssec:lossseg}\addcontentsline{toc}{subsubsection}{\nameref{sssec:lossseg}}
The segmentation loss $\mathcal{L}_s$ is a weighted sum of the focal loss $\mathcal{L}_\mathcal{F}$ \cite{Lin2017} and S{\o}rensen-Dice loss $\mathcal{L}_D$ \cite{Prangemeier2020b,Carion2020}
\begin{equation*}
\mathcal{L}_{s}\left(s_{\sigma\left(i\right)}, \hat{s}_{i}\right) = \lambda_{{\mathcal{F}}}\, \mathcal{L}_{\mathcal{F}}\left(s_{\sigma\left(i\right)}, \hat{s}_{i};\gamma\right)+\lambda_{{\mathcal{D}}}\, \mathcal{L}_{\mathcal{D}}\left(s_{\sigma\left(i\right)}, \hat{s}_{i};\epsilon\right). 
\end{equation*}
The respective weights are $\lambda{_\mathcal{F}}=0.05$ and $\lambda_\mathcal{D}=1$, with focusing parameter $\gamma = 2$
 and $\epsilon = 1$ for numerical stability.

\subsection{Evaluation and implementation}
\label{ssec:eval_imp}
\fakesubsubsection{\P \ Evaluation metrics segmentation (\checkmark)}
\label{sssec:metrics_seg}\addcontentsline{toc}{subsubsection}{\nameref{sssec:metrics_seg}}
We employ a number of metrics to quantitatively analyse the performance of the trained networks with regard to classification, bounding box and segmentation performance. Given the ground truth $\mathbf{Y}$ and the prediction $\mathbf{\hat{Y}}$ (in the corresponding instance-matched permutation), we evaluate the segmentation performance with variants of the Jaccard index $\mathcal{J}$ and the S{\o}rensen-Dice $\mathcal{D}$ coefficient \cite{Prangemeier2020b,Carion2020}, omitting the background
\begin{equation}\label{eqn:J}
\mathcal{D}(\mathbf{Y},\mathbf{\hat{Y}}) = \frac{2 |\mathbf{Y} \cap \mathbf{\hat{Y}}|}{|\mathbf{Y}| + |\mathbf{\hat{Y}}|};  \quad\mathcal{J}_k(\mathbf{Y}_k,\mathbf{\hat{Y}}_k) = \frac{|\mathbf{Y}_k \cap \mathbf{\hat{Y}}_k|}{|\mathbf{Y}_k \cup \mathbf{\hat{Y}}_k|},
\end{equation}
with $\mathcal{J}_k$ intuitively the \textit{intersection-over-union} for each class $k$.
With respect to the metrological application in image cytometry, the cell class is of most importance, therefore, we consider the Jaccard index for the cell class alone ($\mathcal{J}_c$). Similarly, in the case of instance segmentation, we compute $\mathcal{J}_i$ for each instance $i$ and average over all $I$ object instances to compute the mean instance Jaccard index $\bar{\mathcal{J}}_I = \frac{1}{I} \sum_{i = 1}^{I} \mathcal{J}_i$.
\fakesubsubsection{\P \ Evaluation metrics bbox and class (\checkmark)}
\label{sssec:metrics_bbclass}\addcontentsline{toc}{subsubsection}{\nameref{sssec:metrics_bbclass}} 
We utilise the accuracy as the proportion of correct predictions for classification. The bounding boxes are evaluated with the Jaccard index $\bar{\mathcal{J}}_b$. It is defined analogously to the object instance Jaccard index (compare Eqn. \ref{eqn:J}), yet computed implicitly with the bounding box coordinates. 
\fakesubsubsection{\P \ SoA networks: U-Net and Mask R-CNN (\checkmark)}
\label{sssec:SoA_architectures}\addcontentsline{toc}{subsubsection}{\nameref{sssec:SoA_architectures}}
We compare the proposed method with our own implementations of both the state-of-the-art for the trapped yeast application (U-Net \cite{Prangemeier2020b}), as well as more generally with a state-of-the-art instance segmentation meta algorithm (Mask R-CNN \cite{He2017}). The multiclass U-Net for semantic segmentation was implemented  in PyTorch, with the architecture, pre- and post-processing described in \cite{Prangemeier2020b}. We implemented a Mask R-CNN \cite{He2017} with Torchvision (PyTorch) and a ResNet-18 \cite{He2016} backbone, which was pre-trained for image classification.
\fakesubsubsection{\P \ Implementation C-DETR(\checkmark)}
\label{sssec:implementation}
\addcontentsline{toc}{subsubsection}{\nameref{sssec:implementation}}
We implemented the proposed Cell-DETR A and B architectures with PyTorch. We used the MMDetection toolkit \cite{Chen2019} for deformable convolutions and the PyTorch/Cuda implementation for the Pad\'{e} activation units \cite{Molina2019}. We trained the models using AdamW \cite{Loshchilov2018} for optimisation with a weight decay of $10^{-6}$. The initial learning rate was $10^{-5}$ for the backbone and $10^{-4}$ for the rest of the model. The learning rates were decreased by an order of magnitude after $50$ and again $100$ epochs of the total $200$ epochs. The additional first and second-order momentum moving average factors were $0.9$ and $0.999$ respectively. We selected the best performing model based on the cell class Jaccard index $\mathcal{J}_c$, typically after $80$ to $140$ epochs with mini batch size $8$. The training data was randomly augmented by elastic deformation {\cite{Ronneberger2015,Prangemeier2020b}}, horizontal flipping or by the addition of noise with a probability of $0.6$. Inference runtimes for one forward pass were averaged over $1000$ runs on a Nvidia RTX 2080 Ti for all three methods (U-Net, Mask R-CNN and Cell-DETR). 
\subsection{Data acquisition setup }
\label{ssec:acquisition}
\addcontentsline{tdo}{todo}{\textbf{Methodology: Data Acquisition}}
\fakesubsubsection{\P \ Microfluidic culture (\checkmark)}
\label{sssec:mufluidics}
\addcontentsline{toc}{subsubsection}{\nameref{sssec:mufluidics}}
Yeast cells were cultured in a tightly controlled microfluidic environment. A temperature of \SI{30}{\celsius} and the flow of yeast growth media enables yeast to grow for prolonged periods and over multiple cell-cycles. The microfluidic chips confined the cells to the focal plane of the microscope. Continuous media flow hydrodynamically traps the living cells in the microstructures. The Polydimethylsiloxane (PDMS) microstructures constrain the cells in XY, while axial constraints in Z are provided by the cover slip and PDMS ceiling. The space between cover slip and the PDMS ceiling is on the order of a cell diameter to facilitate continuously uniform focus of the cells. 
\fakesubsubsection{\P \ Microscopy (\checkmark)}
\label{sssec:muscopy}
\addcontentsline{toc}{subsubsection}{\nameref{sssec:muscopy}}
We recored time-lapse brightfield (transmitted light) and fluorescent channel imagery of the budding yeast cells every \SI{10}{\min} with a computer controlled microscope (Nikon Eclipe Ti with XYZ stage; $\mu$Manager; 60x objective). A CoolLED pE-100 and a Lumencor SpectraX light engine illuminated the respective channels, which were captured with a ORCA Flash 4.0 (Hamamatsu) camera. Multiple lateral and axial positions were recorded sequentially at each timestep (Fig. \ref{fig:complex}). 

\section{Results}
\label{sec:results}
\addcontentsline{tdo}{todo}{\textbf{Results}}
\subsection{Cell-DETR variant results}
\label{ssec:res_test}
\begin{figure}[htbp]
	\centerline{\includegraphics{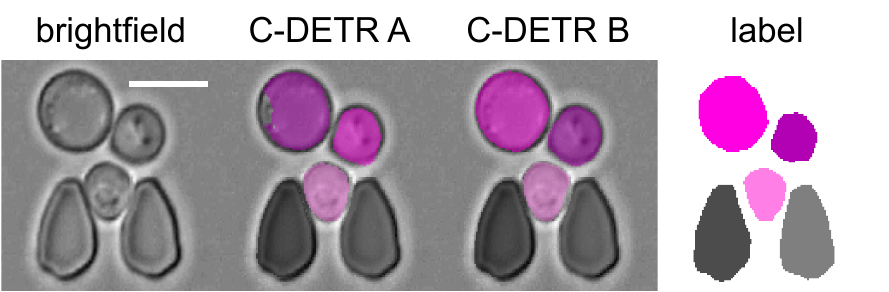}}
		\vspace{-2mm}
	\caption[$\blacksquare$ Fig. \ref{fig:res_AB}:  segmentation examples (\checkmark)]{			Qualitative comparison of Cell-DETR A and B segmentation examples for a selected test image (left) and label (right);
	trap instances in shades of \colorindicator{tab:trap_grey}{dark grey}, cell instances in shades in \colorindicator{tab:detr_violet}{violet}; scale bar \SI{10}{\mu \meter}.}
	\label{fig:res_AB}
\end{figure}

\begin{figure}[htbp]
	\centerline{\includegraphics{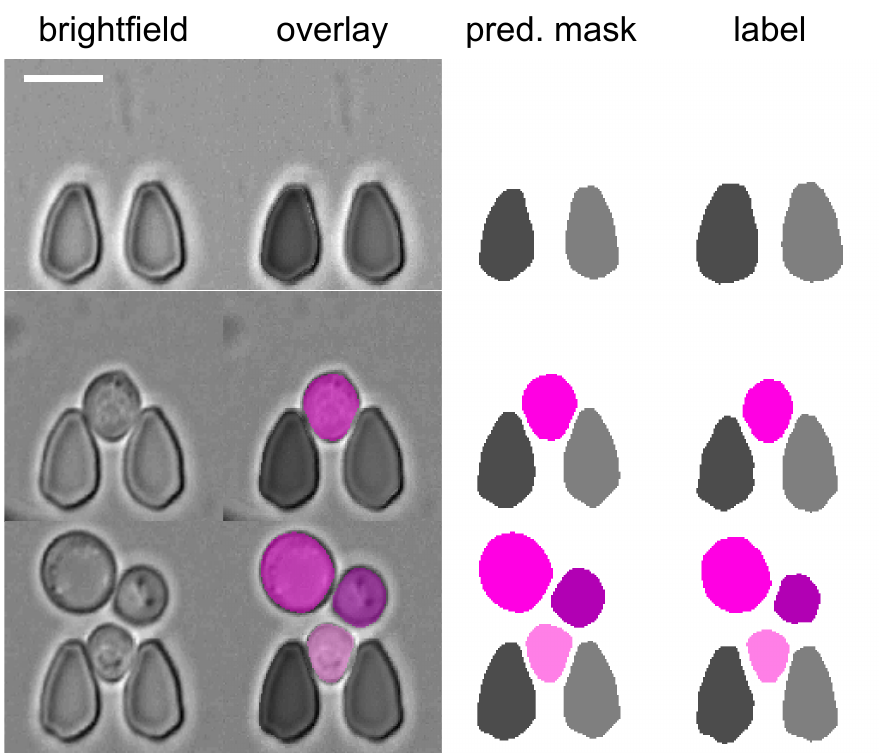}} 
		\vspace{-2mm}
	\caption[$\blacksquare$ Fig. \ref{fig:res_scenarios} test segmentation examples (\checkmark)]{Example of different scenarios from the test dataset segmented with Cell-DETR B: an empty trap (top row), a single trapped cell (middle row) and multiple cells; columns are brightfield, an overlay of the prediction, the prediction mask and the ground truth label  (left to right respectively). Colours indicate traps in shades of \colorindicator{tab:trap_grey}{grey} and cell instances in shades of \colorindicator{tab:detr_violet}{violet}; scale bar \SI{10}{\mu \meter}.}
	\label{fig:res_scenarios}
\end{figure}

\begin{figure}[htbp]
	\centerline{\includegraphics{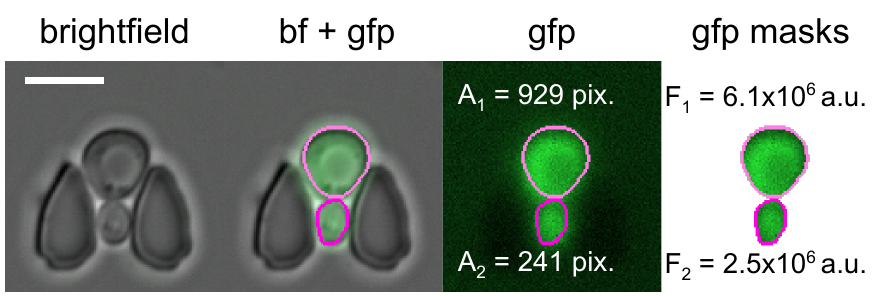}} 
		\vspace{-2mm}
	\caption[$\blacksquare$ Fig. \ref{fig:fluoro}: fluoro seg $\circlearrowleft$]{Example of individual cell fluorescence measurement application with a segmentation mask contour for each individual cell (violet contours \colorindicator{tab:detr_violet}{violet}) based on the brightfield image (left); scale bar \SI{10}{\mu \meter}.}
	\label{fig:fluoro} 
\end{figure}

\addcontentsline{tdo}{subsection}{\nameref{ssec:res_test}}
\fakesubsubsection{\P \ Instance segmentation samples (\checkmark)}
\label{sssec:res_netsamples}
\addcontentsline{toc}{subsubsection}{\nameref{sssec:res_netsamples}}
A sample of segmentation results for the two Cell-DETR variants is shown in Fig. \ref{fig:res_AB}. The cell and trap instances are all detected and classified correctly with slight variations in segmentation contours. Separate instances of cells and traps are indicated by the shades of violet and grey respectively. Variant B demonstrates slightly better segmentation performance. A qualitative example of this is shown in in Fig. \ref{fig:res_AB}, where Cell-DETR A in contrast to B excludes a small section of one cell.
\fakesubsubsection{\P \  Result overview segmentation metrics (\checkmark)}
\label{sssec:res_metrics}
\addcontentsline{toc}{subsubsection}{\nameref{sssec:res_metrics}}
The quantitative comparison between the segmentation performance of the Cell-DETR variants is summarised in Table \ref{tab:detr_seg}. We modified model B for better performance on our application, as described in Section \ref{ssec:DETR_architecture}. The mean Jaccard index over all object instances increased from  $\bar{\mathcal{J}}_I=0.84$ for model A to $\bar{\mathcal{J}}_I=0.85$ for model B, while the cell class Jaccard index increased by a similar margin from $\mathcal{J}_c = 0.83$ to $\mathcal{J}_c = 0.84$. Taking the background into account, a segmentation accuracy of $0.96$ is achieved. Both Cell-DETR surpass the segmentation performance ($\mathcal{J}_c$) of the previous state-of-the-art methods for the trapped yeast application \cite{Bakker2017,Prangemeier2020b}, in addition to directly attaining the instances.  
\begin{table}[htbp]
	\centering
	\caption[$\boxminus$ Table \ref{tab:detr_seg}: DETR segmentation results (\checkmark)]{Segmentation performance of Cell-DETR A and B.}
		\vspace{-2mm}
	\begin{tabularx}{0.95\columnwidth}{l c c c c c c c c} 
		\toprule
		\multicolumn{1}{p{1.5cm}}{\raggedright\textcolor{white}{---------}Model}&%
		\multicolumn{1}{p{1.2cm}}{\centering Sørensen Dice\\ $\mathcal{D}$}&%
		\multicolumn{1}{p{1.2cm}}{\centering Mean instance\\ $\bar{\mathcal{J}}_I$}&%
		\multicolumn{1}{p{1.2cm}}{\centering \textcolor{white}{--------}\\Cell class\\ $\mathcal{J}_{c}$}&%
		\multicolumn{1}{p{1.2cm}}{\centering Seg. \\accuracy }\\
		\midrule
		C-DETR A & 0.92 & 0.84 & 0.83 & 0.96 \\ 
		C-DETR B & 0.92 & 0.85 & 0.84 & 0.96 \\ 
		\bottomrule
	\end{tabularx}
	\label{tab:detr_seg}
\end{table}
\fakesubsubsection{\P \  Result overview bb and classification metrics (\checkmark)}
\label{sssec:res_metrics_bb}
\addcontentsline{toc}{subsubsection}{\nameref{sssec:res_metrics_bb}}
The bounding box and classification performance is summarised in Table \ref{tab:detr_bb_classification}. Again, both models perform similarly well. They correctly classify the object instances (classification accuracy of $1.0$) and detect the correct number of instances for each class. They also perform similarly well at localising the instances, achieve a bounding box \textit{intersection-over-union} of $\mathcal{J}_b = 0.81$, for the standard formulation as well as the generalised form employed for training.
\tds{swap this \P with previous one for better transition to following?}{}
\begin{table}[htbp]
	\centering
	\caption[$\boxminus$ Table \ref{tab:detr_bb_classification}: DETR BB + class results (\checkmark) ]{Bounding box and classification performance metrics for Cell-DETR A and B.
	}
	\vspace{-2mm}
	\begin{tabularx}{0.95\columnwidth}{l c c c c} 
		\toprule
		\multicolumn{1}{p{2.4cm}}{\centering }						&%
		\multicolumn{1}{c}{\centering Bounding box} 				&
		\multicolumn{1}{c}{Classification}							\\  
		\multicolumn{1}{p{2.4cm}}{Model}					&%
		\multicolumn{1}{p{2.4cm}}{\centering Jaccard $\mathcal{J}_b$}	&
		\multicolumn{1}{p{2.4cm}}{\centering accuracy}				\\ 
		\midrule
		C-DETR A & 
		0.81 & 1.0  \\
		C-DETR B & 
		0.81 & 1.0 \\
		\bottomrule
	\end{tabularx}
	\label{tab:detr_bb_classification}
\end{table}
\fakesubsubsection{\P \ Run-time (\checkmark)}
\label{sssec:res_architecture}
\addcontentsline{toc}{subsubsection}{\nameref{sssec:res_architecture}}
The slight increase in segmentation performance that model B yields is a trade off with increased computational cost. The number of parameters is increased from approximately $4 \times 10^6$, to over $5 \times 10^6$ (Table \ref{tab:architecture}). This leads to an increase in runtime from \SI{9.0}{\milli\second} for model A to \SI{21.2}{\milli\second} for model B. These times are orders of magnitude faster than the previous state-of-the-art method DISCO \cite{Bakker2017} and on the same order of magnitude as the currently fastest reported network for this application \cite{Prangemeier2020b}. Runtimes on this order of magnitude suffice for in-the-loop experimental techniques.
\fakesubsubsection{\P \ Example application scenarios  (\checkmark)}
\label{sssec:res_scenarios}
\addcontentsline{toc}{subsubsection}{\nameref{sssec:res_scenarios}}
We select model B for further analysis, based on the improved performance and sufficiently fast runtimes. A selection of segmentation predictions for the three most typical scenarios in the test dataset is given in Fig. \ref{fig:res_scenarios}. The detection of cell and trap instances, without any overlap between instances, is successful for single cells (middle row), multiple cells (bottom row), and empty traps are correctly identified. The introduction of multiple classes (traps, cells), as well as individual object instances facilitated individually segmenting each cell entity and discerning these from both the traps and other cells.
\fakesubsubsection{\P \ Fluorescence measurement (\checkmark)}
\label{sssec:res_fluoro}
\addcontentsline{tdo}{todo}{\textbf{Results: \nameref{sssec:res_fluoro}}}
The intended application of our method is to deliver segmentation masks for each cell instance for subsequent single-cell fluorescence measurements. We trialled this application on unlabelled and unseen data as depicted in Fig. \ref{fig:fluoro}. The cell instances are detected based on the brightfield image (left) and the resulting object segmentation predictions are used as masks to measure the individual cell fluorescence on the fluorescent channel (right). An overlay of the brightfield, fluorescent images with the segmentation contours is depicted in the middle, along with the green fluorescent protein (GFP) channel. The individual cell area ($A_1$ and $A_2$) is measured as the number of pixels in the instance segmentation mask and indicated on the GFP channel. The cell instance fluorescence ($F_1$ and $F_2$) is summed over the mask area and indicated on the right for each individual cell in arbitrary fluorescence units.
\tdsCamera{bf+contour, mask, gfp, gfp x mask}{}
\tdsCamera{make contours thickers, darker, clearer}{}
\subsection{Comparison with state-of-the-art methods} 
\label{ssec:res_comp}
\addcontentsline{tdo}{todo}{\textbf{Results: \nameref{ssec:res_comp}}}
\fakesubsubsection{\P \  Qualitative SoA comparison (\checkmark)}
\label{sssec:res_soaqual}
\addcontentsline{toc}{subsubsection}{\nameref{sssec:res_soaqual}}
We compare our proposed method with the state-of-the-art for the trapped yeast application (DISCO \cite{Bakker2017}, U-Net \cite{Prangemeier2020b}), as well as with a general state-of-the-art method for instance segmentation (Mask R-CNN \cite{He2017}). We implemented both the U-Net and Mask R-CNN methods in this study (Section \ref{ssec:eval_imp}). A characteristic qualitative example of the results is given in Fig. \ref{fig:res_comp_qual}, with the ground truth on the left, followed by Cell-DETR B, Mask R-CNN and U-Net segmentations results. All three methods segment two trap microstructures and all four cells in separate classes, without any overlap or touching cells. Cell-DETR B and Mask R-CNN additionally segment each cell or trap object as an individual instance. The contours are slightly smaller for the U-Net, which is deemed a result of the emphasis on avoiding touching cells and the associated difficulty of discerning these in subsequent post-processing.
\begin{figure}[htbp]
	\centerline{\includegraphics{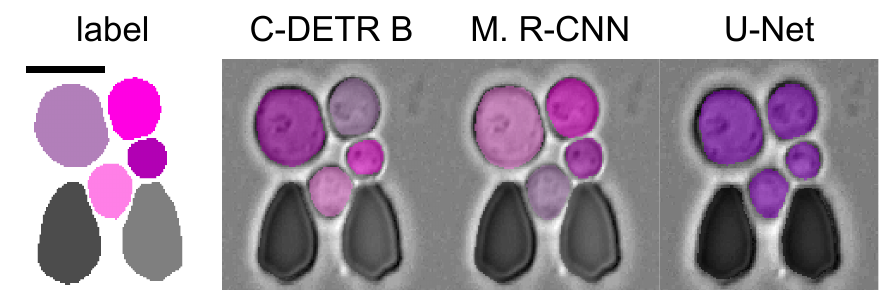}}
		\vspace{-2mm}
	\caption[$\blacksquare$ Fig. \ref{fig:res_comp_qual}:  segmentation examples (\checkmark)]{Example segmentation for our implementations of Cell-DETR B, Mask R-CNN and U-Net.  Trap instances in shades of \colorindicator{tab:trap_grey}{grey} and cell instances in shades of \colorindicator{tab:detr_violet}{violet} (no instance detection for U-Net); scale bar \SI{10}{\mu \meter}.}
	\label{fig:res_comp_qual}
\label{fig:res_comp_qual}
\end{figure}
\fakesubsubsection{\P \ Quantitative SoA comparison, Jacaard (\checkmark)}
\label{sssec:res_soaquant}
\addcontentsline{toc}{subsubsection}{\nameref{sssec:res_soaquant}}
Accurate segmentation of the cells is particularly important for the measurement of cell morphology or fluorescence. We compare the cell class Jaccard index $\mathcal{J}_c$ of our proposed methods Cell-DETR A and B with the application state-of-the-art methods DISCO, U-Net and Mask R-CNN. The comparison is summarised in Table \ref{tab:comparison_sota}. U-Net recently superseded DISCO \cite{Bakker2017} ($\mathcal{J}_c \sim0.7$) as the state-of-the-art trapped yeast segmentation method, achieving $\mathcal{J}_c = 0.82$. Our Cell-DETR variants both further improve on this result, with model B achieving the same $\mathcal{J}_c=0.84$ on par with our Mask R-CNN implementation. Cell-DETR and Mask R-CNN additionally provide each cell object instance.
\begin{table}[!htbp]
	\centering
		\caption[$\boxminus$ Table \ref{tab:comparison_sota}: comparison SotA 	\checkmark]{Comparison of Cell-DETR performance with the state-of-the-art methods for the trapped yeast application (DISCO, U-Net) and instance segmentation (Mask R-CNN).  
	}
	\vspace{-2mm}

	\begin{threeparttable}[t]
			\centering
	\begin{tabular}{l r r c} 
		\toprule
		\multicolumn{1}{p{1.5cm}}{\raggedright \textcolor{white}{-----------} Model}&%
		\multicolumn{1}{p{1.3cm}}{\raggedleft \ \ Cell \ \ \ \\ \ \  Class $\mathcal{J}_{c}$}&%
		\multicolumn{1}{p{1.8cm}}{\raggedleft Inference runtime\tnote{1} \ \
		}&%
		\multicolumn{1}{p{1.0cm}}{\centering \textcolor{white}{------}Instances}\\ 
		\midrule
		DISCO \cite{Bakker2017}\tnote{2} & $\sim 0.70$ & 
		 $\sim \SI{1300}{\milli\second}$ & \texttimes \\
		U-Net & $0.82$ 
		& $\SI{1.8}{\milli\second}
		$& \texttimes \\
		Mask R-CNN & $0.84$ 
		& $\SI{29.8}{\milli\second}
		$ & \checkmark \\
		\textit{Cell-DETR A} & $0.83$ 
		& $\SI{9.0}{\milli\second}
		$ & \checkmark \\
				\textit{Cell-DETR B} & $0.84$ 
		& $\SI{21.2}{\milli\second}
		$ & \checkmark \\
		\bottomrule\\ 
	\end{tabular}
	\vspace{-0.25cm}
\begin{tablenotes}
	\item[1] Runtimes for U-Net, Mask R-CNN, and Cell-DETR averaged over 1000 runs ($\sim300$ different images) on a Nvidia RTX 2080 Ti;
	measurement uncertainty is below $\pm 5\%$.
	\item[2] Reported literature values \cite{Bakker2017}. 
\end{tablenotes}
\end{threeparttable}%
	\label{tab:comparison_sota}
\end{table}
\fakesubsubsection{\P \ Quantitative SoA comparison, run time (\checkmark)}
\label{sssec:res_soatime}
\addcontentsline{toc}{subsubsection}{\nameref{sssec:res_soatime}}
We measured the average runtime of a forward pass of each method on a single specimen image (Table \ref{tab:comparison_sota}). For DISCO \cite{Bakker2017} we consider the reported values that include some pre- and post-processing steps to detect cells individually. The deep methods are significantly faster than DISCO, making online monitoring of live experiments feasible. The U-Net is the fastest, taking \SI{1.8}{\milli\second} for a forward pass, in contrast to \SI{29.8}{\milli\second} for the Mask R-CNN \cite{He2017}. However, the U-Net requires further post-processing steps to detect the object instances and has been reported to take approximately \SI{20}{\milli\second} in conjunction with watershed post-processing \cite{Prangemeier2020b}. The Cell-DETR variants take the middle ground with \SI{9.0}{\milli\second} and \SI{21.2}{\milli\second}.

\section{Discussion}
\label{sec:discussion}
\addcontentsline{tdo}{todo}{\textbf{Discussion}}
\subsection{Analysis of the instance segmentation performance}
\label{ssec:disc_analysis}
\begin{figure}[htbp]
	\newlength\height
	\newlength\width
	\setlength{\height}{5.5cm}
	\setlength{\width}{\columnwidth}
	\centerline{\input{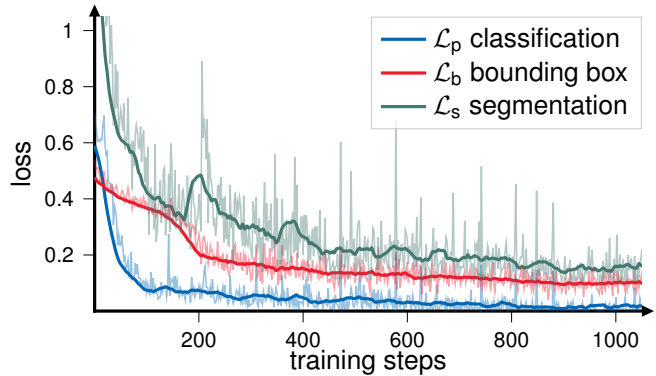}}
	\vspace{-4mm}
	\caption[$\blacksquare$ Fig. \ref{fig:trloss} (\checkmark): training loss plot ]{Classification, bounding boxes and segmentation loss curves for Cell-DETR B; thick lines are running averages (window size 30). 
	}
	\label{fig:trloss}
\end{figure}

\addcontentsline{tdo}{subsection}{\nameref{ssec:disc_analysis}}
\fakesubsubsection{\P \ methody discussion \P}
\label{sssec:disc_limits}
\addcontentsline{toc}{subsubsection}{\nameref{sssec:disc_limits}}
Cell-DETR has some benefits in comparison to state-of-the-art methods, such as Mask R-CNN. The Cell-DETR architecture is comparatively simple and avoids common hand designed components of Mask R-CNNs, such as non-maximum suppression and ROI pooling. This reduces Cell-DETR's reliance on hyperparameters and facilitates end-to-end training with a single combined loss function. In contrast, Mask R-CNNs require additional supervision to train the region proposal network. As a result of these differences, Cell-DETR is easier to implement, has less parameters and is faster than Mask R-CNN for the same segmentation performance.
\fakesubsubsection{\P \ training loss (\checkmark)}
\label{sssec:disc_training}
\addcontentsline{toc}{subsubsection}{\nameref{sssec:disc_training}}
While Cell-DETR does not rely on explicit region proposals, it does utilise attention maps that highlight the pertinent features in the latent space. The mapping of these is learnt during the end-to-end training. The loss curves of individual prediction tasks are shown in Fig. \ref{fig:trloss}. The classification loss $\mathcal{L}_p$ (blue) converges first, indicating that the network first learns how many objects are present in an image and to which class they belong. The bounding box loss $\mathcal{L}_b$ (red) converges next, with the network learning the approximate location of each object. Finally, the model learns to refine the pixel-wise segmentation maps with the segmentation loss $\mathcal{L}_s$ (green) converging last. 

\fakesubsubsection{\P \ Robust, no human intervention (\checkmark)}
\label{sssec:disc_robust}
\addcontentsline{toc}{subsubsection}{\nameref{sssec:disc_robust}}
With respect to the specific single-cell measurement application, Cell-DETR offers robust and repeatable instance segmentation of yeast cells in microstructures. The key cell class segmentation performance surpasses the previous state-of-the-art semantic segmentation methods \cite{Bakker2017, Prangemeier2020b} with a cell class Jaccard index of $0.84$. Additionally, the proposed technique directly detects individual object instances and classifies the objects robustly (near \SI{100}{\percent} accuracy). The robust instance segmentation performance promises to facilitate cell tracking, increase the experimental information yield and enables Cell-DETR to be employed without human intervention.
\subsection{Limitations, outlook and future potential} 
\label{ssec:disc_outlook}
\fakesubsubsection{\P \ methody outlookish \P}
\label{sssec:disc_limits}
\addcontentsline{toc}{subsubsection}{\nameref{sssec:disc_limits}}
\fakesubsubsection{\P \ limitations: extend dataaset classes etc (\checkmark)}
\label{sssec:disc_limits}
\addcontentsline{toc}{subsubsection}{\nameref{sssec:disc_limits}}
The presented models are trained for a specific microfluidic configuration and trap geometry. While they are relatively robust and fulfil their intended purpose, their utility could be broadened by expanding the dataset to include more classes, for example different trap geometries. More generally as an instance segmentation method, Cell-DETR offers a platform for incorporating future advances in attention mechanisms as they are increasingly outperforming convolutional approaches. For example, replacing the convolutional elements in the backbone and segmentation head with axial-attention \cite{Wang2020} may lead to further improved performance. Currently, Cell-DETR achieves state-of-the-art performance and as an instance segmentation method is generally suitable for and readily adaptable to a wide range of biomedical imaging applications. 

\fakesubsubsection{\P \ future potential online pipeline (\checkmark)}
\label{sssec:disc_pot}
\addcontentsline{toc}{subsubsection}{\nameref{sssec:disc_pot}}
The presented Cell-DETR methods can be harnessed for high-content quantitative single-cell TLFM. Cell-DETR, Mask R-CNN and U-Net achieve runtimes orders of magnitudes faster than the previous state-of-the-art trapped yeast method (DISCO \cite{Bakker2017}).  These runtimes coupled with Cell-DETRs robust instance segmentation make both online monitoring and closed-loop optimal experimental design of typical experiments with approximately $1000$ traps feasible. Harnessing this potential promises to provide increased experimental information yields and greater biological insights in the future. 


\section{Conclusion}
\label{sec:conclusion}
\addcontentsline{tdo}{todo}{\textbf{Conclusion - rough}}
In summary, we present Cell-DETR, an attention-based transformer method for direct instance segmentation and showcase it on a typical application. To the best of our knowledge, this is the first application of detection transformers on biomedical data. The proposed method has fewer parameters and is $30 \%$ faster while matching the segmentation performance of a state-of-the-art Mask R-CNN. A simpler Cell-DETR variant exhibits slightly lesser segmentation performance ($\mathcal{J}_c = 0.83$ instead of $0.84$) while requiring $1/3$rd of a Mask R-CNN's runtime. As a general instance segmentation model, Cell-DETR achieves state-of-the-art performance and is deemed suitable and readily adaptable for a range of biomedical imaging applications.

Showcased on a typical systems or synthetic biology application, the proposed Cell-DETR robustly detects each cell instance and directly provides instance-wise segmentation maps suitable for cell morphology and fluorescence measurements. In comparison to the previous semantic segmentation trapped yeast baselines, Cell-DETR provides better segmentation performance with a cell class Jaccard index $\mathcal{J}_c = 0.84$ while additionally detecting each individual cell instance and maintaining comparable runtimes. This promises to reduce measurement uncertainty, facilitate cell tracking efficacy and increase the experimental data yield in future applications. The resulting runtimes and accurate instance segmentation make future online monitoring feasible, for example for closed-loop optimal experimental control.

\vspace{8mm}
\section*{Acknowledgements}
\label{sec:acknowledgements}
\addcontentsline{tdo}{todo}{\textbf{sec:acknowledgements}}
We thank Christian Wildner for insightful discussions, Andr\'e O. Fran\c{c}ani and Jan Basrawi for contributing to labelling and Markus Baier for aid with the computational setup.
\indent
This work was supported by the Landesoffensive f\"{u}r wissenschaftliche Exzellenz as part of the LOEWE Schwerpunkt CompuGene. H.K. acknowledges support from the European Research Council (ERC) with the  consolidator grant CONSYN (nr. 773196).
\vspace{8mm}

\bibliographystyle{IEEEtran}
\bibliography{IEEEabrv,detr_bib}

\newpage

\end{document}